%% file: main.tex
\documentclass[10pt,twocolumn,letterpaper]{article}

\usepackage{iccv}
\usepackage{times}
\usepackage{epsfig}
\usepackage{graphicx}
\usepackage{amsmath}
\usepackage{amssymb}
\usepackage{comment}
\usepackage{subcaption}
\usepackage{stfloats}
\usepackage[pagebackref=true,breaklinks=true,letterpaper=true,colorlinks,bookmarks=false]{hyperref}

\iccvfinalcopy

\newcommand{\ANNCOUNT}{5102\xspace}
\newcommand{\CLASSCOUNT}{31\xspace}
\newcommand{\SCANOBJCOUNT}{3979\xspace}
\newcommand{\CADTOTALCOUNT}{7650\xspace}

\newcommand{\DATASET}{Scan-CAD Object Similarity}
\newcommand{\IMPROVEMENT}{$12\%$\xspace}

\def\iccvPaperID{}

\ificcvfinal\fi
\begin{document}

\title{Joint Embedding of 3D Scan and CAD Objects}

\author{
Manuel Dahnert$^{1}$ \qquad Angela Dai$^{1}$ \qquad Leonidas Guibas$^{2,3}$ \qquad Matthias Nie{\ss}ner$^{1}$
\vspace{0.2cm} \\ 
$^{1}$Technical University of Munich \qquad $^{2}$Stanford University \qquad $^{3}$Facebook AI Research
}

\twocolumn[{%
	\renewcommand\twocolumn[1][]{#1}%
	\maketitle
	\begin{center}
 		\vspace{-0.3cm}
		\includegraphics[width=0.95\textwidth]{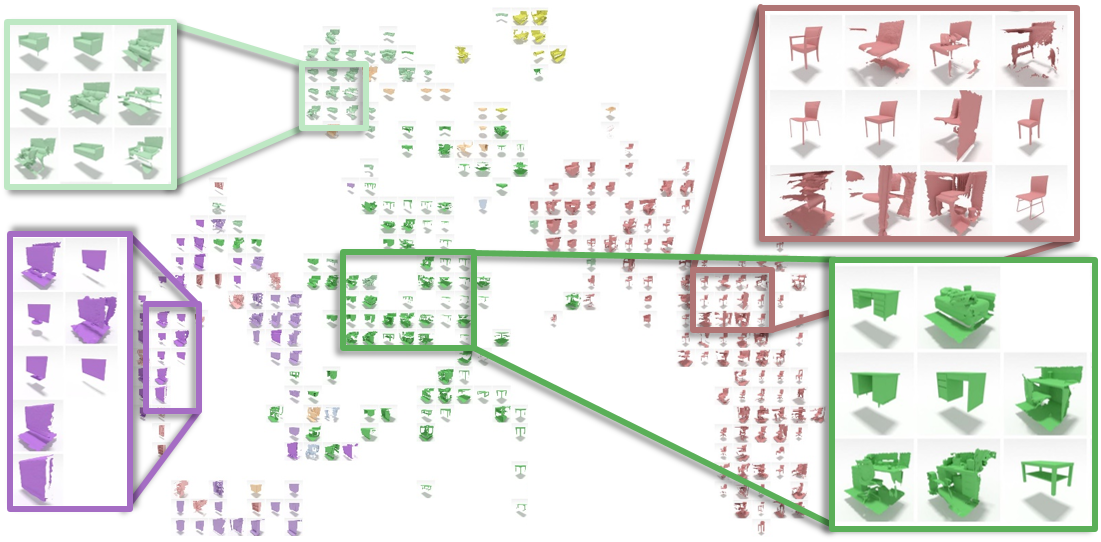}
	 \end{center}
	 	\vspace{-0.7cm}
	    \captionof{figure}{We learn a joint embedding space of scan and CAD object geometry, visualized here by t-SNE. Semantically similar objects lie close together, despite very different lower-level geometric characteristics (clutter, noise, partialness, etc).}
    	\label{fig:teaser}
	\vspace{0.3cm}
}]

\input{0abstract.tex}
\input{1introduction.tex}
\input{2relatedwork.tex}

\input{3overview.tex}

\input{4network.tex}

\input{5dataset.tex}
\input{6results.tex}

\input{7conclusion.tex}

\section*{Acknowledgements}
This work is supported by Occipital, the ERC Starting Grant Scan2CAD (804724), a Google Faculty Award, the ZD.B., ONR MURI grant N00014-13-1-0341, NSF grant IIS-1763268, as well as a Vannevar Bush Fellowship.
We would also like to thank the support of the TUM-IAS, funded by the German Excellence Initiative and the European Union Seventh Framework Programme under grant agreement n° 291763, for the TUM-IAS Rudolf M{\"o}{\ss}bauer Fellowship and Hans Fischer Senior Faculty Fellowship (Focus Group Visual Computing).

{\small
\bibliographystyle{ieee_fullname}
\bibliography{main}
}

\newpage
\input{appendix.tex}

\end{document}

%% file: 0abstract.tex
\begin{abstract}
3D scan geometry and CAD models often contain complementary information towards understanding environments, which could be leveraged through establishing a mapping between the two domains.
However, this is a challenging task due to strong, lower-level differences between scan and CAD geometry.
We propose a novel approach to learn a joint embedding space between scan and CAD geometry, where semantically similar objects from both domains lie close together.
To achieve this, we introduce a new 3D CNN-based approach to learn a joint embedding space representing object similarities across these domains.
To learn a shared space where scan objects and CAD models can interlace, we propose a stacked hourglass approach to separate foreground and background from a scan object, and transform it to a complete, CAD-like representation to produce a shared embedding space.
This embedding space can then be used for CAD model retrieval; to further enable this task, we introduce a new dataset of ranked scan-CAD similarity annotations, enabling new, fine-grained evaluation of CAD model retrieval to cluttered, noisy, partial scans.
Our learned joint embedding outperforms current state of the art for CAD model retrieval by \IMPROVEMENT{} in instance retrieval accuracy.
\end{abstract}

%% file: 1introduction.tex
\section{Introduction}
\label{sec:intro}

The capture and reconstruction of real-world 3D scenes has seen significant progress in recent years, driven by increasing availability of commodity RGB-D sensors such as the Microsoft Kinect or Intel RealSense.
State-of-the-art 3D reconstruction approaches can achieve impressive reconstruction fidelity with robust tracking~\cite{newcombe2011kinectfusion,izadi2011kinectfusion,niessner2013hashing,whelan2015elasticfusion,choi2015robust,dai2017bundlefusion}.
Such 3D reconstructions have now begun to drive forward 3D scene understanding with the recent availability of annotated reconstruction datasets~\cite{dai2017scannet,Matterport3D}.
With the simultaneous availability of synthetic CAD model datasets~\cite{shapenet2015}, we have an opportunity to drive forward both 3D scene understanding and geometric reconstruction.

3D models of scanned real-world objects as well as synthetic CAD models of shapes contain significant information about understanding environments, often in a complementary fashion. 
Where CAD models often comprise relatively simple, clean, compact geometry, real-world objects are often more complex, and scanned real-world object geometry is then more complex, as well as noisy and incomplete.
It is thus very informative to establish mappings between the two domains -- for instance, to visually transform scans to CAD representations, or transfer learned semantic knowledge from CAD models to a real-world scan.
Such a semantic mapping is difficult to obtain due to the lack of exact matches between synthetic models and real-world objects and these strong, low-level geometric differences.

Current approaches towards retrieving CAD models representative of scanned objects thus focus on the task of retrieving a CAD model of the correct object class category~\cite{qi2016volumetric,dai2017scannet,huashrec2017,phamshrec2018}, without considering within-class similarities or rankings.
In contrast, our approach learns a joint embedding space of scan and CAD object geometry where similar objects from both domains lie close together as shown in Fig.~\ref{fig:teaser}.
To this end, we introduce a new 3D CNN based approach to learn a semantically mixed embedding space as well as a dataset of \ANNCOUNT scan-CAD ranked similarity annotations.
Using this dataset of scan-CAD similarity, we can now fully evaluate CAD model retrieval, with benchmark evaluation of retrieval accuracy as well as ranking ability.
To learn a joint embedding space, our model takes a stacked hourglass approach of a series of encoder-decoders: first learning to disentangle a scan object from its background clutter, then transforming the partial scan object to a complete object geometry, and finally learning a shared embedding with CAD models through a triplet loss.
This enables scan and CAD object geometry into a shared space and outperforms state-of-the-art CAD model retrieval approaches by \IMPROVEMENT{} in instance retrieval accuracy.

In summary, we make the following contributions:
\begin{itemize}
    \item We propose a novel stacked hourglass approach leveraging a triplet loss to learn a joint embedding space between CAD models and scan object geometry.
    \item We introduce a new dataset of ranked scan-CAD object similarities, establishing a benchmark for CAD model retrieval from an input scan object.
    For this task, we propose fine-grained evaluation scores for both retrieval and ranking.
\end{itemize}

%% file: 2relatedwork.tex
\section{Related Work}
\label{sec:relatedWork}

\paragraph{3D Shape Descriptors}
Characterizations of 3D shapes by compact feature descriptors enable a variety of tasks in shape analysis such as shape matching, retrieval, or organization.
Shape descriptors have thus seen a long history in geometry processing.
Descriptors for characterizing 3D shapes have been proposed leveraging handcrafted features based on lower-level geometric characteristics such as volume, distance, or curvature~\cite{osada2002shape,ohbuchi2003shape,gal2007pose,rusu2009fast,tombari2010signature}, or higher-level characteristics such as topology~\cite{hilaga2001topology,chen20023d,sundar2003skeleton}.
Characterizations in the form of 2D projections of the 3D shapes have also been proposed to describe the appearance and geometry of a shape~\cite{chen2003visual}.
Recently, with advances in deep neural networks for 3D data, neural networks trained for point cloud or volumetric shape classification have also been used to provide feature descriptors for 3D shapes \cite{qi2016volumetric,qi2017pointnet}.

\paragraph{CAD Model Retrieval for 3D Scans}
CAD model retrieval to RGB-D scan data has been increasingly studied with the recent availability of large-scale datasets of real-world~\cite{dai2017scannet,Matterport3D} and synthetic~\cite{shapenet2015} 3D objects.
The SHREC challenges~\cite{huashrec2017,phamshrec2018} for CAD model retrieval to real-world scans of objects have become very popular in this context.
Due to lack of ground truth data for similarity of CAD models to scan objects, CAD model retrieval in this context is commonly evaluated using the class categories as a coarse proxy for similarity; that is, a retrieved model is considered to be a correct retrieval if the category matches that of the query scan object.
We propose a finer-grained evaluation for the task of CAD model retrieval for a scan object with our new \DATASET{} dataset and benchmark.

\paragraph{Multi-modal Embeddings}
Embedding spaces across different data modalities have been used for various computer vision tasks, such as establishing relationships between image and language~\cite{weston2010large,weston2011wsabie}, or learning similarity between different image domains such as photos and product images~\cite{bell2015learning}.
These cross-domain relationships have been shown to aid tasks such as object detection~\cite{peng2015learning, massa2016deep}.
More recently, Herzog et al. proposed an approach to relate 3D models, keywords, images, and sketches~\cite{herzog2015lesss}.
Li et al. also introduced a CNN-based approach to learn a shared embedding space between CAD models and images, leveraging a CNN to map images into a pre-constructed feature space of CAD model similarity~\cite{li2015jointembedding}.
Our approach also leverages a CNN to construct a model which can learn a joint embedding between scan objects and CAD models in an end-to-end fashion, learning to become invariant to differences in partialness or geometric noise.

%% file: 3overview.tex
\begin{figure*}[tbp]
    \centering
    \includegraphics[width=0.92\textwidth]{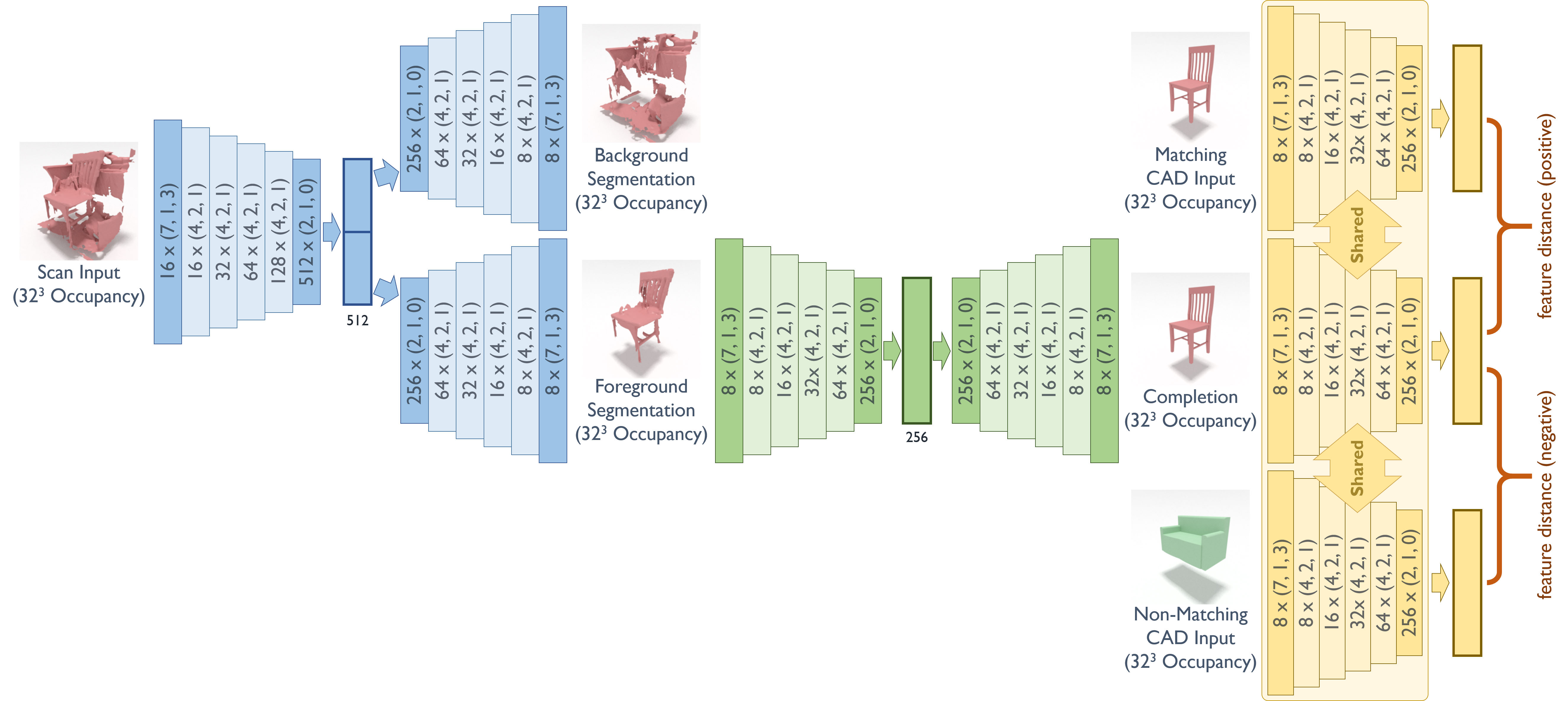}
    \vspace{-0.3cm}
    \caption{Our network architecture to construct a joint embedding between scan and CAD object geometry.
    The architecture is designed in a stacked hourglass fashion, with a series of hourglass encoder-decoders to transform a scan input to a more CAD-like representation, before mapping the features into an embedding space with a triplet loss. 
    The first hourglass (blue) segments a scan object from its background clutter, the second hourglass (green) predicts the complete geometry for the segmented object, from which the final feature encoding is computed (yellow); CAD object features are computed with the same final encoder.
    Note that layers are denoted with parameters $c\times (k,s,p)$ with number of output channels $c$, kernel size $k$, stride $s$, and padding $p$. Lighter colored layers denote residual blocks, darker colored layers denote a convolutional layer.
    \vspace{-0.2cm}}
    \label{fig:architecture}
\end{figure*}

\section{Method Overview}
\label{sec:overview}

Our method learns a shared embedding space between real-world scans of objects and CAD models, where semantically similar scan and CAD objects lie near each other, with scan and CAD objects mixed together, invariant to lower-level geometric differences (partialness, noise, etc).

We represent both scan and CAD objects by binary grids representing voxel occupancy, and design a 3D convolutional neural network to encourage scan objects and CAD objects to map into a shared embedding space.
Our model is thus structured in a stacked hourglass~\cite{newell2016stacked} fashion, designed to transform scan objects to a more CAD-like representation before mapping them into this joint space.

The first hourglass learns to segment the scan geometry into object and background clutter, using an encoder with two decoders trained to reconstruct foreground and background, respectively.
The segmented foreground then leads to the next hourglass, composed of an encoder-decoder trained to reconstruct the complete geometry of the segmented but partial scan object.
This helps to disentangle confounding factors like clutter and partialness of scanned objects before mapping them into a shared space with CAD objects.
Here, the completed scan is then input to an encoder to train a latent feature vector which maps into this embedding space, by constraining the latent space to match that of a CAD encoder on a matching CAD object and be far away from the encoder for a non-matching CAD object.

This enables learning of a joint embedding space where semantically similar CAD objects and scan objects lie mixed together.
With this learned shared embedding space, we can enable applications such as much finer-grained CAD model retrieval to scan objects than previously attainable.
To this end, we demonstrate our joint scan-CAD embedding in the context of CAD model retrieval, introducing a \DATASET{} benchmark and evaluation scores for this task.

%% file: 4network.tex
\section{Learning a Joint Scan-CAD Embedding}
\label{sec:embedding}

\subsection{Network Architecture}

Our network architecture is shown in Fig.~\ref{fig:architecture}.
It is an end-to-end, fully-convolutional 3D neural network designed to disentangle lower-level geometric differences between scan objects and CAD models.
During training, we take as input a scan object $S$ along with a corresponding CAD model $C_p$ and a dissimilar CAD model $C_n$, each represented by its binary occupancy in a $32^3$ volumetric grid.
At test time, we use the learned feature extractors for scan or CAD objects to compute a feature vector in the joint embedding space.

The model is composed as a stacked hourglass of two encoder-decoders  followed by a final encoder.
The first two hourglass components focus on transforming a scan object to a more CAD-like representation to encourage the joint embedding space to focus on higher-level semantic and structural similarities between scan and CAD than lower-level geometric differences.

The first hourglass is thus designed to segment a scan object from nearby background clutter (e.g., floor, wall, other objects), and is composed of an encoder and two decoders (one for the foreground scan object, one for the background).
The encoder employs an initial convolution followed by a series of $4$ residual blocks, and a final convolution layer resulting in a $512$-dimensional latent feature space.
This feature is then split in half; the first half feeds into a decoder which reconstructs the segmented scan object from background, and the second half to a decoder which reconstructs the background clutter of the input scan geometry.
The decoders are structured symmetrically to the encoder (each using half the feature channels).
For predicted scan object geometry $x_{fg}$ and background geometry $x_{bg}$, we train with a proxy loss $\mathcal{L}_{seg} = BCE(x_{fg}, \textrm{gt}_{fg}) + BCE(x_{bg}, \textrm{gt}_{bg})$ for reconstructing segmented scan object and background clutter, respectively, as occupancy grids.

The second hourglass takes the segmented scan object and aims to generate the complete geometry of the object, as real-world scans often result in partially observed geometry.
This is structured in encoder-decoder fashion, where the encoder and decoder are structured symmetrically to the decoders of the first segmentation hourglass.
We then employ a proxy loss on the completion as an occupancy grid: $\mathcal{L}_{cmp} = BCE(x_{cmp}, C_p)$, for completion prediction $x_{cmp}$ and CAD model $C_p$ corresponding to the scan object.

The final encoder aims to learn the joint scan-CAD embedding space.
This is formulated as a triplet loss:
\begin{equation*}
    \mathcal{L} = \max ( d( f(S), g(C_p) ) - d( f(S), g(C_n) ) + \textrm{margin}, 0),
\end{equation*}
where $f(S) = f^e(f^c(f^s(S)))$ with $f^s$ representing the scan segmentation, $f^c$ the scan completion, and $f^e$ an encoder structured symmetrically to the encoder of $f^c$ which produces a feature vector of size $256$.
$g(C)$ is an encoder structured identically to $f^e$ which computes the feature vector for a CAD occupancy grid. 
For all our experiments, all losses are weighted equally and we use Euclidean distance and a margin of $0.2$.

\subsection{Network Training}

We train our model end-to-end from scratch.
For training data, we use the paired scan and CAD models ($S$ and $C_p$), from  Scan2CAD~\cite{avetisyan2019scan2cad}, which provides CAD model alignments from ShapeNet~\cite{shapenet2015} onto the real-world scans of ScanNet~\cite{dai2017scannet}.
For the non-matching CAD models $C_n$, we randomly sample models from Scan2CAD from different class categories. After every epoch we re-sample new negatives.

We train our model using an Adam optimizer with a batch size of 128 and an initial learning rate of $0.001$, which is decayed by $10$ every $20k$ iterations.
Our model is trained for $100k$ iterations ($\approx$ 1 day) on a single Nvidia GTX 1080Ti.

%% file: 5dataset.tex
\begin{table*}[t]
    \centering
     \resizebox{\textwidth}{!}{
\begin{tabular}{|l|rrrrrrrrrr|r|rr|}
\hline
Method & trash bin & bathtub & bed & bookshelf & cabinet & chair & display & file & sofa & table & {\bf class avg} & {\bf inst (k=10)} & {\bf inst (k=50)} \\
\hline\hline
FPFH~\cite{rusu2009fast}      & 0.09	& 0.06 & 0.01 & 0.03 & 0.02 & 0.05 & 0.08 & 0.02 & 0.02 & 0.01 & 0.03 & 0.02 & 0.04 \\
SHOT~\cite{tombari2010signature}      & 0.17 & 0.14 & 0.06 & 0.02 & 0.03 & 0.12 & 0.13 & 0.08 & 0.01 & 0.05 & 0.08 & 0.04 & 0.07 \\
PointNet~\cite{qi2017pointnet}    & 0.10 & 0.08 & 0.18 & 0.08 & 0.03 & 0.07 & 0.06 & 0.12 & 0.04 & 0.05 & 0.06 & 0.05 & 0.13\\
3DCNN~\cite{qi2016volumetric}      & 0.29 & 0.31 & 0.32 & 0.31 & 0.21 & 0.14 & 0.29 & 0.28 & 0.29 & 0.18 & 0.22 & 0.20 & 0.33 \\ \hline
Ours (no seg, no cmpl)             & 0.14 & 0.13 & 0.23 & 0.11 & 0.07 & 0.15 & 0.14 & 0.28 & 0.19 & 0.18 & 0.16 & 0.14 & 0.22 \\
Ours (no cmpl)  & 0.24 & 0.32 & 0.26 & 0.28 & 0.13 & 0.21 & 0.44 & 0.24 & 0.19 & 0.25 & 0.24 & 0.21 & 0.31 \\
Ours (no seg)  & \textbf{0.50} & 0.53 & 0.52 & \textbf{0.51} & 0.48 & 0.44 & 0.51 & 0.53 & 0.47 & \textbf{0.50} & 0.49 & 0.48 & 0.49 \\
Ours (no triplet) & 0.51 & \textbf{0.48} & 0.45 & 0.22 & 0.42 & 0.34 & 0.25 & \textbf{0.50} & 0.28 & 0.38 & 0.36 & 0.34 & 0.42 \\
Ours (w/o end-to-end) & 0.42 & 0.46 & 0.46 & 0.35 & 0.42 & 0.35 & 0.33 & 0.51 & 0.34 & 0.41 & 0.39 & 0.37 & 0.44 \\
Ours & 0.51 & \textbf{0.52} & \textbf{0.50} & \textbf{0.51} & \textbf{0.51} & \textbf{0.48} & \textbf{0.50} & 0.55 & \textbf{0.51} & 0.49 & \textbf{0.50} & \textbf{0.49} & \textbf{0.50} \\
\hline
\end{tabular}
}
\vspace{-0.3cm}
\caption{Evaluation of the joint scan-CAD embedding space. 
We compare our learned scan-CAD feature space to those constructed from features computed through both handcrafted and learned shape descriptors. 
We evaluate the confusion between scan and CAD, where $0.5$ reflects a perfect confusion.
}
\label{tab:embedding_confusion}
\end{table*}

\section{\DATASET{} Benchmark}
\label{sec:dataset}

\begin{figure}
\begin{center}
\includegraphics[width=0.9\columnwidth]{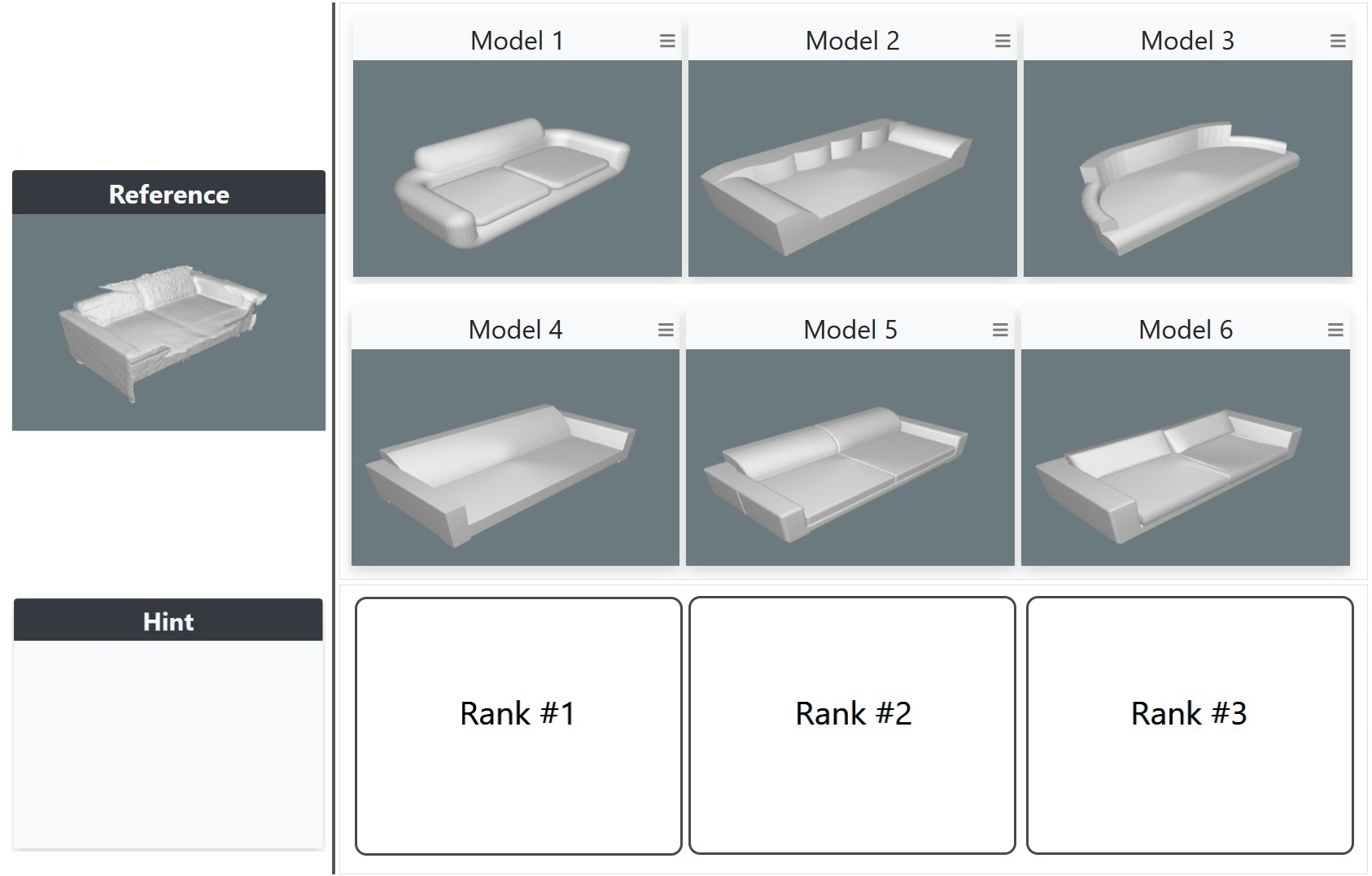}
\end{center}
\vspace{-0.7cm}
   \caption{Annotation interface for obtaining ranked similarity of CAD models to a scan query. A user selects and ranks up to 3 CAD models from a pool of 6 proposed models.
\vspace{-0.4cm}
   }
\label{fig:annotation_interface}
\end{figure}

Our learned joint embedding space between scan and CAD object geometry enables characterization of these objects at higher-level semantic and structural similarity.
This allows us to formulate applications like CAD model retrieval in a more comprehensive fashion, in particular in contrast to previous approaches which evaluate retrieval by the class accuracy of the retrieved object~\cite{huashrec2017,qi2016volumetric,phamshrec2018}.
We aim to characterize retrieval through finer-grained object similarity than class categories.
Thus, we propose a new \DATASET{} dataset and benchmark for CAD model retrieval.

To construct our \DATASET{} dataset, we develop an intuitive annotation web interface designed to measure scan-CAD similarities, inspired by \cite{liu2015style}.
As shown in Fig.~\ref{fig:annotation_interface}, the geometry of a query scan model is shown, along with a set of $6$ CAD models.
A user then selects up to $3$ similar CAD models from the proposed set, in order of similarity to the query scan geometry, resulting in ranked scan-CAD similarity annotations.
Users are instructed to measure the similarity in terms of object geometry. Initially, the models are displayed in a canonical pose, but the user can rotate, translate or zoom each model individually to inspect it in closer detail.
As scan objects can occasionally be very partial, we also provide an option to click on a `hint' which shows a color image of the object with a bounding box around it, in order to help identify the object if the segmented geometry is insufficient.

To collect these scan-CAD similarity annotations, we use segmented scan objects from the ScanNet dataset~\cite{dai2017scannet}, which provides labeled semantic instance segmentation over the scan geometry. 
CAD models are proposed from ShapeNetCore~\cite{shapenet2015}.
The CAD model proposals are sampled leveraging the annotations from the Scan2CAD dataset~\cite{avetisyan2019scan2cad}, which provides CAD model alignments for $3049$ unique ShapeNetCore models to objects in $1506$ ScanNet scans.
We propose CAD models for a scan query by sampling in the latent space of an autoencoder trained on ShapeNetCore using the feature vector of the associated CAD model from the Scan2CAD dataset. In the latent space, we select the 30 nearest neighbors of the associated CAD model and randomly select $6$ to be proposed to the user. 
This enables a description of ranked similarity for a scan object to several CAD models, which we can then use for fine-grained evaluation of CAD model retrieval.

\paragraph{Dataset Statistics}
To construct our \DATASET{} dataset and benchmark, we employed three university students as annotators, and trained them to become familiar with the interface and to ensure high-quality annotations for our task.
Our final dataset is composed of \ANNCOUNT annotations covering \CLASSCOUNT different class categories (derived from ShapeNet classes).
These cover \SCANOBJCOUNT unique scan objects and \CADTOTALCOUNT unique CAD models.

\begin{table*}[bp]
    \centering
     \resizebox{\textwidth}{!}{
\begin{tabular}{|l|rrrrrrrrrrr|rr|}
\hline
Method & trash bin & bathtub & bed & bookshelf & cabinet & chair & display & file & sofa & table & other & {\bf class avg} & {\bf inst avg} \\
\hline\hline
FPFH~\cite{rusu2009fast}      & 0.02 & 0.07 & 0.00 & 0.00 & 0.00 & 0.18 & 0.03 & 0.00 & 0.07 & 0.02 & 0.03 & 0.04 & 0.08 \\
SHOT~\cite{tombari2010signature}      & 0.00 & 0.20 & 0.09 & 0.00 & 0.01 & 0.06 & 0.12 & 0.00 & 0.07 & 0.02 & 0.03 & 0.05 & 0.04 \\
PointNet~\cite{qi2017pointnet}    & 0.38 & 0.00 & \textbf{0.61} & 0.23 & 0.04 & 0.43 & 0.37 & 0.17 & 0.09 & 0.13 & 0.07 & 0.23 & 0.29 \\
3DCNN~\cite{qi2016volumetric}       & \textbf{0.52} & 0.33 & 0.48 & \textbf{0.46} & 0.14 & 0.28 & 0.38 & \textbf{0.33} & 0.17 & 0.18 & 0.32 & 0.33 & 0.31 \\
\hline
Ours (no seg, no cmpl)        & 0.06 & 0.00 & 0.15 & 0.04 & 0.00 & 0.47 & 0.30 & 0.00 & 0.20 & 0.13 & 0.04 & 0.13 & 0.23 \\
Ours (no cmpl)        & 0.13 & 0.07 & 0.15 & 0.12 & 0.04 & 0.37 & 0.38 & 0.00 & 0.15 & 0.26 & 0.09 & 0.16 & 0.24 \\
Ours (no seg)        & 0.14 & 0.07 & 0.24 & 0.13 & 0.15 & 0.40 & 0.32 & 0.17 & 0.15 & 0.21 & 0.13 & 0.19 & 0.26 \\
Ours (no triplet)        & 0.03 & 0.13 & 0.39 & 0.04 & 0.11 & 0.07 & 0.08 & 0.00 & 0.13 & 0.09 & 0.04 & 0.10 & 0.08 \\
Ours (w/o end-to-end)         & 0.42 & 0.27 & 0.48 & 0.07 & 0.15 & 0.42 & 0.27 & 0.25 & \textbf{0.35} & 0.21 & 0.32 & 0.29 & 0.32 \\
Ours        & 0.50 & \textbf{0.60} & 0.42 & 0.19 & \textbf{0.26} & \textbf{0.55} & \textbf{0.45} & 0.25 & 0.33 & \textbf{0.32} & \textbf{0.43} & \textbf{0.39} & \textbf{0.43} \\
\hline
\end{tabular}
}
\vspace{-0.3cm}
\caption{Top-1 retrieval accuracy for CAD model retrieval on the test split of our \DATASET{} benchmark.
\vspace{-0.2cm} }
\label{tab:retrieval_comparison}
\end{table*}

\subsection{Benchmark Evaluation}\label{subsec:benchmark_eval}
We also introduce a new benchmark to evaluate both a scan-CAD embedding space as well as CAD model retrieval.
To evaluate the learned embedding space, we measure a \emph{confusion} score: for each object embedding feature, we compute the percentage of scan neighbors and the percentage of CAD neighbors for its $k$ nearest neighbors.
The final confusion score is then 
\begin{equation*}
    \resizebox{\columnwidth}{!}{
$0.5\left(\frac{1}{k|{\textrm{scans}}|}\sum_{\textrm{scans}}|\{\textrm{CAD nbrs}\}| + \frac{1}{k|{\textrm{cads}}|}\sum_{\textrm{CADs}}|\{\textrm{scan nbrs}\}|\right)$.
    }
\end{equation*}
This describes how well the embedding space mixes the two domains, agnostic to the lower-level geometric differences. 
Note that we evaluate this confusion score on a set of embedded scan and CAD features with a 1-to-1 mapping between the scan and CAD objects, and use $k=10$. A confusion of $0.5$ means a perfect balance between scan and CAD objects in the local neighborhood around an object.

To evaluate the semantic embedding quality, we propose two scores for scan-CAD similarity in the context of CAD model retrieval: \emph{retrieval accuracy} and \emph{ranking quality}.
Here, we employ the scan-CAD similarity annotations of our \DATASET{} dataset.
For both retrieval accuracy and ranking quality, we consider an input query scan, and retrieval from the set of $6$ proposed CAD models supplemented with $100$ additional randomly selected CAD models of different class from the query (in order to reflect a diverse set of models for retrieval).
For retrieval accuracy, we evaluate whether the top-$1$ retrieved model lies in the set of models annotated as similar to the query scan.
We also evaluate the ranking; that is, for a ground truth annotation with $n$ rank-annotated similar models ($n\leq 3$), we take the top $n$ predicted models and evaluate the number of models predicted in the correct rank divided by $n$.

Note that for the task of CAD model retrieval, we consider scan objects in the context of potential background clutter from scanning; that is we assume a given object detection as input, but not object segmentation.

%% file: 6results.tex
\begin{figure*}[tbp]
    \centering
    \includegraphics[width=0.9\textwidth]{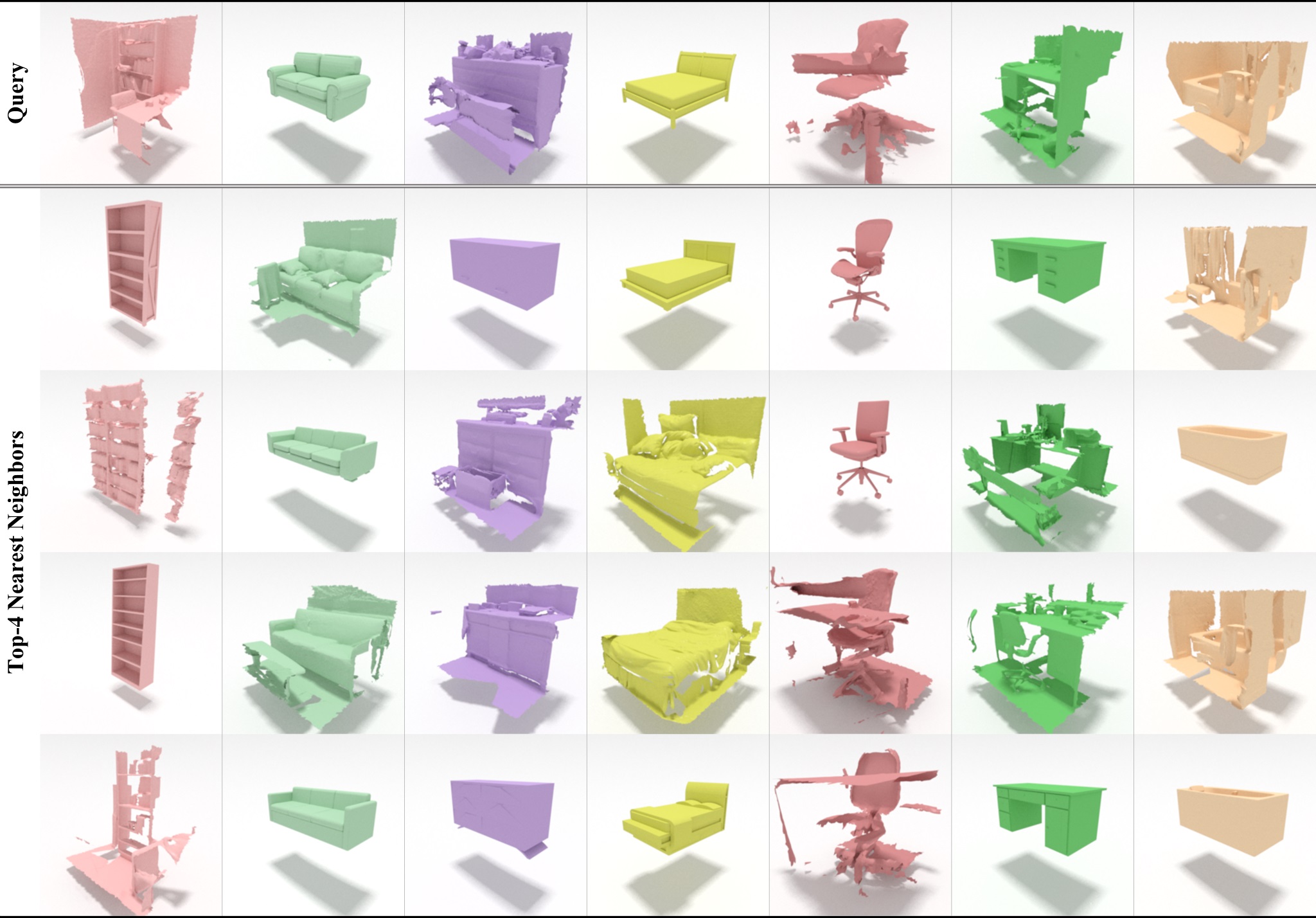}
\vspace{-0.3cm}
    \caption{Our CAD model retrieval results, visualizing the top $4$ retrieved models using our joint embedding space for various scan and CAD queries. Our feature space learns to mix together scan and CAD objects in a semantically meaningful fashion.
\vspace{-0.3cm}}
    \label{fig:ours_nn5_retrieval}
\end{figure*}

\begin{figure*}[tbp]
    \centering
    \includegraphics[width=0.9\textwidth]{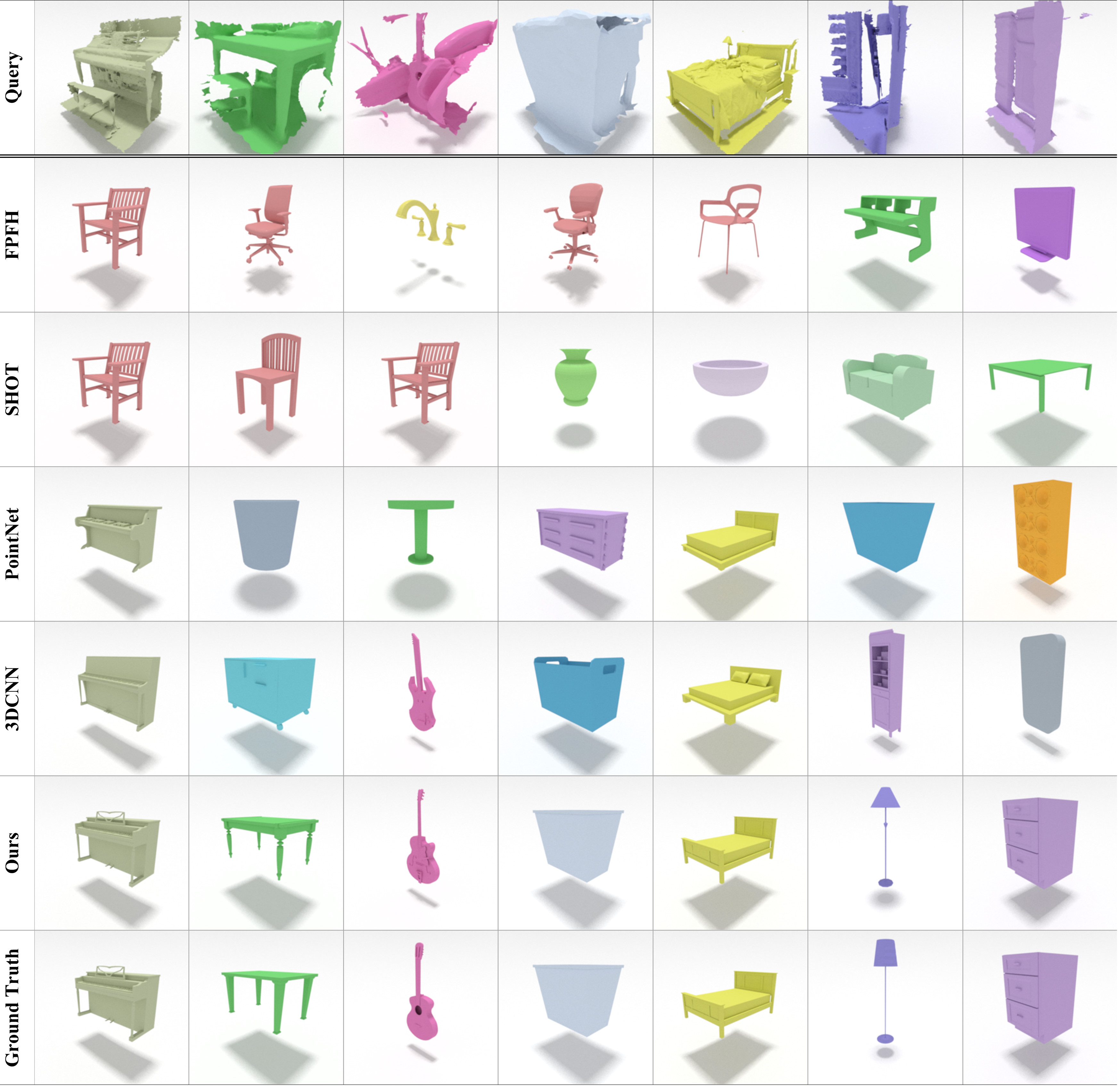}
    \vspace{-0.3cm}
    \caption{CAD model retrieval results (top-1) for various scan queries (from left to right: piano, table, guitar, trash bin, bed, lamp, dresser).
    Our approach to a joint embedding of scan and CAD can retrieve similar models at a finer-grained level than state-of-the-art handcrafted (FPFH~\cite{rusu2009fast}, SHOT~\cite{tombari2010signature}) and learned (PointNet~\cite{qi2017pointnet}, 3DCNN~\cite{qi2016volumetric}) 3D object descriptors.
    \vspace{-0.2cm}
    }
    \label{fig:comparison_retrieval}
\end{figure*}

\section{Results and Evaluation}
\label{sec:results}

We evaluate both the quality of our learned scan-CAD embedding space as well as its application to the task of CAD model retrieval for scan objects using the \emph{confusion}, \emph{retrieval accuracy}, and \emph{ranking quality} scores proposed in Section~\ref{subsec:benchmark_eval}.
Additionally, in Table~\ref{tab:classbased_retrievalmetric}, we evaluate on a coarser level retrieval score based on whether the retrieved model's class is correct, which is the basis of retrieval evaluation used in previous approaches~\cite{qi2016volumetric,huashrec2017,phamshrec2018}.
We compare our method with both state-of-the-art handcrafted shape descriptors FPFH~\cite{rusu2009fast} and SHOT~\cite{tombari2010signature} as well as learned shape descriptors from PointNet~\cite{qi2017pointnet} and the volumetric 3D CNN from \cite{qi2016volumetric}.
We evaluate FPFH and SHOT on point clouds uniformly sampled from the mesh surface of the scans and CAD objects, with all meshes normalized to lie within a unit sphere. We compute a single shape descriptor for the entire object by using the centroid of the mesh and a radius of 1. 

We train PointNet on $1024$ points uniformly sampled from the scan and CAD objects for object classification, and extract the $256$-dimensional feature vector before the final classification layer. 
For the volumetric 3D CNN of \cite{qi2016volumetric}, we train on $32^3$ occupancy grids of both scan objects and CAD models, and extract the $512$-dimensional feature vector before the final  classification layer.

\begin{figure*}[bp]
    \centering
    \includegraphics[width=\textwidth]{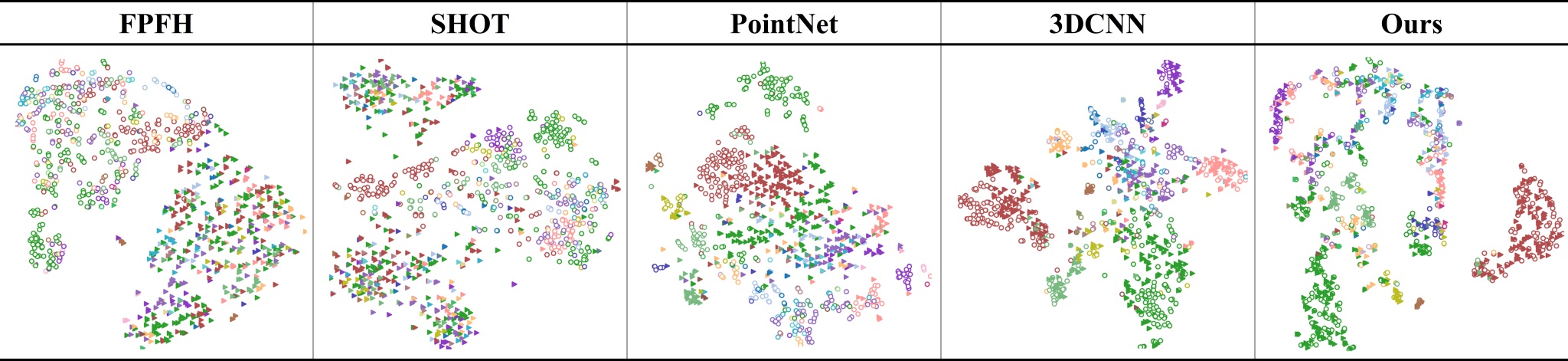}
\vspace{-0.7cm}
    \caption{Comparison of latent spaces visualized by t-SNE. Filled triangles represent scan objects, circles represent CAD models.
    While FPFH, SHOT, and PointNet result in almost entirely disjoint clusters, 3DCNN is able to co-locate the classes of both domains next to each other, but does not confuse them.
    Our approach learns an embedding space where scan and CAD objects mix together but remain semantically structured.}
    \label{fig:tsne_comparison}
\end{figure*}

\paragraph{Learned joint embedding space.}
In Table~\ref{tab:embedding_confusion}, we show that our model is capable of learning a very mixed embedding space, where scan and CAD objects lie about as close to each other as they do to other objects from the same domain, while maintaining semantic structure in the space. 
In contrast, both previous handcrafted and learned shape descriptors result in segregated feature spaces with scan objects lying much closer to scan than CAD objects and vice versa, see Fig.~\ref{fig:tsne_comparison}.
Our learned scan-CAD embedding space is shown in Fig.~\ref{fig:teaser}, visualized by t-SNE.
We also show the top-4 nearest neighbors for various queries from our established joint embedding space in  Fig.~\ref{fig:ours_nn5_retrieval}, retrieving objects from both domains while maintaining semantic structure.

\paragraph{Comparison to alternative CAD model retrieval approaches.}
Using our learned feature embedding space for scan and CAD objects, we evaluate it for the task of CAD model retrieval to scan object geometry.
Tables~\ref{tab:retrieval_comparison} and \ref{tab:ranking_comparison} show our CAD retrieval quality in comparison to alternative 3D object descriptors, using our benchmark evaluation. Fig.~\ref{fig:comparison_retrieval} shows the top-1 CAD retrievals  for various scan queries. 
Our learned features from the joint embedding space achieve notably improved retrieval on both a class accuracy-based retrieval score (Table~\ref{tab:classbased_retrievalmetric}) as well as our proposed finer-grained retrieval evaluation scores.

\begin{table}[t]
    \centering
     \resizebox{0.45\columnwidth}{!}{
\begin{tabular}{|l|r|r|}
\hline
  Method  & Top-1 & Top-5 \\
\hline\hline
FPFH~\cite{rusu2009fast} & 0.14 & 0.13 \\
SHOT~\cite{tombari2010signature} & 0.07  & 0.08 \\
PointNet~\cite{qi2017pointnet} &  0.49 &  0.45 \\
3DCNN~\cite{qi2016volumetric} &  0.57 & 0.47 \\
\hline
Ours & \textbf{0.68} & \textbf{0.62} \\
\hline
\end{tabular}
}
\vspace{-0.2cm}
\caption{Evaluation of CAD model retrieval by Top-1 and Top-5 using category-based evaluation of retrieval accuracy.
\vspace{-0.4cm}}
\label{tab:classbased_retrievalmetric}
\end{table}

\paragraph{How much do the segmentation and completion steps matter?}
Tables~\ref{tab:embedding_confusion}, \ref{tab:retrieval_comparison}, and \ref{tab:ranking_comparison} show that the proxy segmentation and completion steps in transforming scan object geometry to a more CAD-like representation are important towards learning an effective joint embedding space as well as for CAD model retrieval, with performance improving by $20\%$ and $23\%$ with segmentation and completion, respectively, for our retrieval accuracy (class average).
Additionally, we show that end-to-end training significantly improves the learned embedding space.

\begin{table*}[tp]
    \centering
     \resizebox{\textwidth}{!}{
\begin{tabular}{|l|rrrrrrrrrrr|rr|}
\hline
Method & trash bin & bathtub & bed & bookshelf & cabinet & chair & display & file & sofa & table & other & {\bf class avg} & {\bf inst avg}  \\
\hline\hline
FPFH~\cite{rusu2009fast}      & 0.01 & 0.09 & 0.01 & 0.00 & 0.00 & 0.06 & 0.01 & 0.00 & 0.03 & 0.01 & 0.02 & 0.02 & 0.03 \\
SHOT~\cite{tombari2010signature}      & 0.00 & 0.06 & 0.01 & 0.00 & 0.01 & 0.03 & 0.02 & 0.00 & 0.04 & 0.01 & 0.01 & 0.02 & 0.02 \\
PointNet~\cite{qi2017pointnet}    & 0.22 & 0.03 & 0.24 & 0.15 & 0.04 & 0.16 & 0.11 & 0.00 & 0.02 & 0.04 & 0.05 & 0.10 & 0.12 \\
3DCNN~\cite{qi2016volumetric}       & 0.23 & 0.03 & \textbf{0.31} & \textbf{0.16} & 0.07 & 0.11 & 0.12 & 0.13 & 0.09 & 0.07 & \textbf{0.12} & 0.12 & 0.13 \\
\hline
Ours (no seg, no cmpl)        & 0.05 & 0.00 & 0.08 & 0.03 & 0.01 & 0.17 & \textbf{0.14} & 0.00 & 0.10 & 0.04 & 0.04 & 0.06 & 0.10 \\
Ours (no cmpl)        & 0.08 & 0.00 & 0.06 & 0.04 & 0.02 & 0.15 & 0.12 & 0.06 & 0.06 & \textbf{0.11} & 0.05 & 0.07 & 0.10 \\
Ours (no seg)        & 0.08 & 0.06 & 0.12 & 0.08 & 0.09 & 0.14 & 0.09 & 0.06 & 0.07 & 0.07 & 0.04 & 0.08 & 0.10 \\
Ours (no triplet)        & 0.01 & 0.06 & 0.13 & 0.03 & 0.04 & 0.03 & 0.02 & 0.06 & 0.04 & 0.04 & 0.05 & 0.05 & 0.04 \\
Ours (w/o end-to-end)         & 0.14 & 0.18 & 0.12 & 0.04 & 0.06 & 0.18 & \textbf{0.14} & 0.13 & \textbf{0.16} & 0.08 & \textbf{0.12} & 0.12 & 0.13 \\
Ours        & \textbf{0.29} & \textbf{0.24} & 0.19 & 0.08 & \textbf{0.12} & \textbf{0.19} & \textbf{0.14} & \textbf{0.19} & 0.15 & 0.10 & 0.09 & \textbf{0.16} & \textbf{0.16}  \\
\hline
\end{tabular}
}
\vspace{-0.3cm}
\caption{Ranking quality of CAD model retrieval on the test split of our \DATASET{} benchmark. 
\vspace{-0.2cm}}
\label{tab:ranking_comparison}
\end{table*}

\paragraph{What is the impact of the triplet loss formulation?}
Using a triplet loss to train the feature embedding in a shared space significantly improves the construction of the embedding space, as well as CAD model retrieval from the space.
In Tables~\ref{tab:embedding_confusion}, \ref{tab:retrieval_comparison}, and \ref{tab:ranking_comparison}, we show a comparison to training our model using only positive scan-CAD associations rather than both positive and negative samples; the triplet constraint of both positive and negative examples produces a much more globally structured embedding space.

\paragraph{How robust is the model to rotations?}
To achieve robustness to rotations for scan queries, we can train our method with rotation augmentation, achieving similar performance for arbitrarily rotated scan inputs (0.42 instance average retrieval accuracy, 0.16 instance average ranking quality). See the appendix for more detail.

%% file: 7conclusion.tex
\subsection{Limitations}

While our approach learns an effective embedding space between scan and CAD object geometry, there are still several important limitations.
For instance, we only consider the geometry of the objects in both scan and CAD domain; considering color information would potentially be another powerful signal for joint embedding or CAD model retrieval.
The geometry is also represented as an occupancy grid, which can limit resolution of fine detail.
For the CAD model retrieval task, we currently assume a given object detection, and while 3D object detection has recently made significant progress, detection and retrieval would likely benefit from an end-to-end formulation.

\section{Conclusion}
\label{sec:conclusion}

In this paper, we have presented a 3D CNN-based approach to jointly map scan and CAD object geometry into a shared embedding space.
Our approach leverages a stacked hourglass architecture combined with a triplet loss to transform scan object geometry to a more CAD-like representation, effectively learning a joint feature embedding space.
We show the advantages of our learned feature space for the task of CAD model retrieval, and propose several new evaluation scores for finer-grained retrieval evaluation, with our approach outperforming state-of-the-art handcrafted and learned methods on all evaluation scores.
We hope that learning such a joint scan-CAD embedding space will not only open new possibilities for CAD model retrieval but also potentially enable further perspective on mapping or reciprocal transfer of knowledge between the two domains.

%% file: appendix.tex
\begin{appendix}

\section{Additional Quantitative Studies}

We provide several additional quantitative experiments evaluating robustness against rotation, as well as the performance of the object segmentation and completion from our stacked hourglass model.

\paragraph{Robustness to rotations}
To achiever robustness to potential rotations for input scan queries, we train our method with rotation augmentation around the up axis. 
During training, we rotate the initial partial and cluttered scan object as well as the positive and negative CAD model with the same random rotation. 
During evaluation of CAD model retrieval, we embed CAD models into the embedding space using $12$ uniform rotations for each CAD model; for an input scan query, we then find the closest CAD embedding.
We train for 160k iterations and a triplet margin of 0.1. 
Table~\ref{tab:rotation} shows the results of CAD model retrieval while testing on randomly rotated scan object inputs.
With this rotation augmentation, we can achieve performance on par with the case of canonically oriented objects while testing on arbitrarily rotated scan inputs: 0.42 in instance average retrieval accuracy and 0.16 in instance average ranking quality.

\begin{table}[h]
    \centering
     \resizebox{\columnwidth}{!}{
\begin{tabular}{|l|r|}
\hline
Method & \textbf{IoU} \\
\hline\hline
[A] SGPN~\cite{wang2018sgpn}     &  0.10 \\

[B] Segmentation(Ours)     &  \textbf{0.36} \\
\hline
\hline
[C] Segmentation(Ours) + 3D-EPN~\cite{dai2017complete} & 0.48 \\

[D] Segmentation(Ours) + Completion(Ours)     &  \textbf{0.53} \\
\hline
\end{tabular}
}
\caption{Evaluation (IoU) of our segmentation and completion to SGPN~\cite{wang2018sgpn} and 3D-EPN~\cite{dai2017complete}, respectively.
}
\label{tab:iou}
\end{table}

\begin{table*}[t]
    \centering
     \resizebox{\textwidth}{!}{
\begin{tabular}{|l|rrrrrrrrrrr|rr|}
\hline
Method & trash bin & bathtub & bed & bookshelf & cabinet & chair & display & file & sofa & table & other & {\bf class avg} & {\bf inst avg}  \\
\hline\hline
Ours with rotations       & 0.35 & 0.47 & 0.45 & 0.18 & 0.30 & 0.56 & 0.45 & 0.08 & 0.41 & 0.41 & 0.26 & 0.36 & 0.42  \\
\hline
\end{tabular}
}
\vspace{-0.2cm}
\caption{Evaluation of CAD model retrieval by retrieval accuracy on our \DATASET{} benchmark. 
\vspace{-0.2cm}}
\label{tab:rotation}
\end{table*}

\paragraph{What is the performance of the segmentation and completion?}
In Table~\ref{tab:iou}, we evaluate the performance of the first and second hourglass with Intersection over Union (IoU) between the predicted and ground truth binary occupancy grid. 
We compare our model against SGPN~\cite{wang2018sgpn}, a point cloud based segmentation method, and 3D-EPN~\cite{dai2017complete}, a voxel-based object completion network.
For evaluation, we then convert all outputs to occupancy grids to compute the final IoU scores.

Additionally, we evaluate our stacked hourglass model, replacing our completion encoder-decoder with the model of 3D-EPN, trained end-to-end to learn a joint scan-CAD embedding.
Tables~\ref{tab:embedding_confusion_supp}, \ref{tab:retrieval_comparison_supp}, and \ref{tab:ranking_comparison_supp} show that our model achieves notably better performance in embedding space confusion as well as CAD model retrieval and ranking than the version using 3D-EPN.

\begin{table*}[t]
    \centering
     \resizebox{\textwidth}{!}{
\begin{tabular}{|l|rrrrrrrrrr|r|rr|}
\hline
Method & trash bin & bathtub & bed & bookshelf & cabinet & chair & display & file & sofa & table & {\bf class avg} & {\bf inst (k=10)} & {\bf inst (k=50)} \\
\hline\hline
Ours (no seg, no cmpl)             & 0.14 & 0.13 & 0.23 & 0.11 & 0.07 & 0.15 & 0.14 & 0.28 & 0.19 & 0.18 & 0.16 & 0.14 & 0.22 \\
Ours (no cmpl)  & 0.24 & 0.32 & 0.26 & 0.28 & 0.13 & 0.21 & 0.44 & 0.24 & 0.19 & 0.25 & 0.24 & 0.21 & 0.31 \\
Ours (no seg)  & \textbf{0.50} & 0.53 & 0.52 & \textbf{0.51} & 0.48 & 0.44 & 0.51 & 0.53 & 0.47 & \textbf{0.50} & 0.49 & 0.48 & 0.49 \\
Ours (no triplet) & 0.51 & 0.48 & 0.45 & 0.22 & 0.42 & 0.34 & 0.25 & \textbf{0.50} & 0.28 & 0.38 & 0.36 & 0.34 & 0.42 \\
Ours (3D-EPN~\cite{dai2017complete} for cmpl) & 0.45 & \textbf{0.51} & 0.49 & 0.46 & \textbf{0.50} & 0.33 & 0.40 & 0.53 & 0.47 & 0.45 & 0.43 & 0.42 & 0.47 \\
Ours (w/o end-to-end) & 0.42 & 0.46 & 0.46 & 0.35 & 0.42 & 0.35 & 0.33 & 0.51 & 0.34 & 0.41 & 0.39 & 0.37 & 0.44 \\
Ours & 0.51 & 0.52 & \textbf{0.50} & \textbf{0.51} & 0.51 & \textbf{0.48} & \textbf{0.50} & 0.55 & \textbf{0.51} & 0.49 & \textbf{0.50} & \textbf{0.49} & \textbf{0.50} \\
\hline
\end{tabular}
}
\vspace{-0.3cm}
\caption{Evaluation of the joint scan-CAD embedding space. 
We compare our learned scan-CAD feature space to those constructed from features computed through both handcrafted and learned shape descriptors. 
We evaluate the confusion between scan and CAD, where $0.5$ reflects a perfect confusion.
}
\label{tab:embedding_confusion_supp}
\end{table*}

\begin{table*}[t]
    \centering
     \resizebox{\textwidth}{!}{
\begin{tabular}{|l|rrrrrrrrrrr|rr|}
\hline
Method & trash bin & bathtub & bed & bookshelf & cabinet & chair & display & file & sofa & table & other & {\bf class avg} & {\bf inst avg} \\
\hline \hline
Ours (no seg, no cmpl)        & 0.06 & 0.00 & 0.15 & 0.04 & 0.00 & 0.47 & 0.30 & 0.00 & 0.20 & 0.13 & 0.04 & 0.13 & 0.23 \\
Ours (no cmpl)        & 0.13 & 0.07 & 0.15 & 0.12 & 0.04 & 0.37 & 0.38 & 0.00 & 0.15 & 0.26 & 0.09 & 0.16 & 0.24 \\
Ours (no seg)        & 0.14 & 0.07 & 0.24 & 0.13 & 0.15 & 0.40 & 0.32 & 0.17 & 0.15 & 0.21 & 0.13 & 0.19 & 0.26 \\
Ours (no triplet)        & 0.03 & 0.13 & 0.39 & 0.04 & 0.11 & 0.07 & 0.08 & 0.00 & 0.13 & 0.09 & 0.04 & 0.10 & 0.08 \\
Ours (3D-EPN~\cite{dai2017complete} for cmpl)         & 0.41 & 0.33 & 0.42 & \textbf{0.21} & 0.19 & 0.49 & 0.40 & 0.08 & 0.20 & 0.30 & 0.31 & 0.30 & 0.37 \\
Ours (w/o end-to-end)         & 0.42 & 0.27 & \textbf{0.48} & 0.07 & 0.15 & 0.42 & 0.27 & \textbf{0.25} & \textbf{0.35} & 0.21 & 0.32 & 0.29 & 0.32 \\
Ours        & \textbf{0.50} & \textbf{0.60} & 0.42 & 0.19 & \textbf{0.26} & \textbf{0.55} & \textbf{0.45} & \textbf{0.25} & 0.33 & \textbf{0.32} & \textbf{0.43} & \textbf{0.39} & \textbf{0.43} \\
\hline
\end{tabular}
}
\vspace{-0.2cm}
\caption{Evaluation of CAD model retrieval by top-1 retrieval accuracy on the test split of  our \DATASET{} benchmark.
\vspace{-0.2cm} }
\label{tab:retrieval_comparison_supp}
\end{table*}

\begin{table*}[t]
    \centering
     \resizebox{\textwidth}{!}{
\begin{tabular}{|l|rrrrrrrrrrr|rr|}
\hline
Method & trash bin & bathtub & bed & bookshelf & cabinet & chair & display & file & sofa & table & other & {\bf class avg} & {\bf inst avg}  \\
\hline\hline
Ours (no seg, no cmpl)        & 0.05 & 0.00 & 0.08 & 0.03 & 0.01 & 0.17 & 0.14 & 0.00 & 0.10 & 0.04 & 0.04 & 0.06 & 0.10 \\
Ours (no cmpl)        & 0.08 & 0.00 & 0.06 & 0.04 & 0.02 & 0.15 & 0.12 & 0.06 & 0.06 & \textbf{0.11} & 0.05 & 0.07 & 0.10 \\
Ours (no seg)        & 0.08 & 0.06 & 0.12 & 0.08 & 0.09 & 0.14 & 0.09 & 0.06 & 0.07 & 0.07 & 0.04 & 0.08 & 0.10 \\
Ours (no triplet)        & 0.01 & 0.06 & 0.13 & 0.03 & 0.04 & 0.03 & 0.02 & 0.06 & 0.04 & 0.04 & 0.05 & 0.05 & 0.04 \\
Ours (3D-EPN~\cite{dai2017complete} for cmpl)         & 0.17 & 0.18 & \textbf{0.19} & \textbf{0.09} & \textbf{0.12} & 0.17 & \textbf{0.15} & 0.06 & 0.10 & \textbf{0.11} & 0.09 & 0.13 & 0.14 \\
Ours (w/o end-to-end)         & 0.14 & 0.18 & 0.12 & 0.04 & 0.06 & 0.18 & 0.14 & 0.13 & \textbf{0.16} & 0.08 & \textbf{0.12} & 0.12 & 0.13 \\
Ours        & \textbf{0.29} & \textbf{0.24} & \textbf{0.19} & 0.08 & \textbf{0.12} & \textbf{0.19} & 0.14 & \textbf{0.19} & 0.15 & 0.10 & 0.09 & \textbf{0.16} & \textbf{0.16}  \\
\hline
\end{tabular}
}
\vspace{-0.2cm}
\caption{Evaluation of CAD model retrieval by ranking quality on the test split of our \DATASET{} benchmark. 
\vspace{-0.2cm}}
\label{tab:ranking_comparison_supp}
\end{table*}

\end{appendix}

%% file: main.bbl
\begin{thebibliography}{10}\itemsep=-1pt

\bibitem{avetisyan2019scan2cad}
Armen Avetisyan, Manuel Dahnert, Angela Dai, Manolis Savva, Angel~X. Chang, and
  Matthias Nie{\ss}ner.
\newblock Scan2cad: Learning cad model alignment in rgb-d scans.
\newblock In {\em Proc. Computer Vision and Pattern Recognition (CVPR), IEEE},
  2019.

\bibitem{bell2015learning}
Sean Bell and Kavita Bala.
\newblock Learning visual similarity for product design with convolutional
  neural networks.
\newblock {\em ACM Transactions on Graphics (TOG)}, 34(4):98, 2015.

\bibitem{Matterport3D}
Angel~X. Chang, Angela Dai, Thomas Funkhouser, Maciej Halber, Matthias
  Niessner, Manolis Savva, Shuran Song, Andy Zeng, and Yinda Zhang.
\newblock {Matterport3D}: Learning from {RGB-D} data in indoor environments.
\newblock {\em International Conference on 3D Vision (3DV)}, 2017.

\bibitem{shapenet2015}
Angel~X. Chang, Thomas Funkhouser, Leonidas Guibas, Pat Hanrahan, Qixing Huang,
  Zimo Li, Silvio Savarese, Manolis Savva, Shuran Song, Hao Su, Jianxiong Xiao,
  Li Yi, and Fisher Yu.
\newblock {ShapeNet: An Information-Rich 3D Model Repository}.
\newblock Technical Report arXiv:1512.03012 [cs.GR], Stanford University ---
  Princeton University --- Toyota Technological Institute at Chicago, 2015.

\bibitem{chen20023d}
Ding-Yun Chen and Ming Ouhyoung.
\newblock A 3d object retrieval system based on multi-resolution reeb graph.
\newblock In {\em Proc. of Computer Graphics Workshop}, volume~16, 2002.

\bibitem{chen2003visual}
Ding-Yun Chen, Xiao-Pei Tian, Yu-Te Shen, and Ming Ouhyoung.
\newblock On visual similarity based 3d model retrieval.
\newblock In {\em Computer graphics forum}, volume~22, pages 223--232. Wiley
  Online Library, 2003.

\bibitem{choi2015robust}
Sungjoon Choi, Qian-Yi Zhou, and Vladlen Koltun.
\newblock Robust reconstruction of indoor scenes.
\newblock In {\em 2015 IEEE Conference on Computer Vision and Pattern
  Recognition (CVPR)}, pages 5556--5565. IEEE, 2015.

\bibitem{dai2017scannet}
Angela Dai, Angel~X. Chang, Manolis Savva, Maciej Halber, Thomas Funkhouser,
  and Matthias Nie{\ss}ner.
\newblock {ScanNet}: Richly-annotated {3D} reconstructions of indoor scenes.
\newblock In {\em Proc. Computer Vision and Pattern Recognition (CVPR), IEEE},
  2017.

\bibitem{dai2017bundlefusion}
Angela Dai, Matthias Nie{\ss}ner, Michael Zollh{\"o}fer, Shahram Izadi, and
  Christian Theobalt.
\newblock Bundlefusion: Real-time globally consistent 3d reconstruction using
  on-the-fly surface reintegration.
\newblock {\em ACM Transactions on Graphics (TOG)}, 36(3):24, 2017.

\bibitem{dai2017complete}
Angela Dai, Charles~Ruizhongtai Qi, and Matthias Nie{\ss}ner.
\newblock Shape completion using 3d-encoder-predictor cnns and shape synthesis.
\newblock In {\em Proc. Computer Vision and Pattern Recognition (CVPR), IEEE},
  2017.

\bibitem{gal2007pose}
Ran Gal, Ariel Shamir, and Daniel Cohen-Or.
\newblock Pose-oblivious shape signature.
\newblock {\em IEEE transactions on visualization and computer graphics},
  13(2):261--271, 2007.

\bibitem{herzog2015lesss}
Robert Herzog, Daniel Mewes, Michael Wand, Leonidas Guibas, and Hans-Peter
  Seidel.
\newblock Lesss: Learned shared semantic spaces for relating multi-modal
  representations of 3d shapes.
\newblock In {\em Computer Graphics Forum}, volume~34, pages 141--151. Wiley
  Online Library, 2015.

\bibitem{hilaga2001topology}
Masaki Hilaga, Yoshihisa Shinagawa, Taku Kohmura, and Tosiyasu~L Kunii.
\newblock Topology matching for fully automatic similarity estimation of 3d
  shapes.
\newblock In {\em Proceedings of the 28th annual conference on Computer
  graphics and interactive techniques}, pages 203--212. ACM, 2001.

\bibitem{huashrec2017}
Binh-Son Hua, Quang-Trung Truong, Minh-Khoi Tran, Quang-Hieu Pham, Asako
  Kanezaki, Tang Lee, HungYueh Chiang, Winston Hsu, Bo Li, Yijuan Lu, et~al.
\newblock Shrec’17: Rgb-d to cad retrieval with objectnn dataset.

\bibitem{izadi2011kinectfusion}
Shahram Izadi, David Kim, Otmar Hilliges, David Molyneaux, Richard Newcombe,
  Pushmeet Kohli, Jamie Shotton, Steve Hodges, Dustin Freeman, Andrew Davison,
  et~al.
\newblock Kinectfusion: real-time 3d reconstruction and interaction using a
  moving depth camera.
\newblock In {\em Proceedings of the 24th annual ACM symposium on User
  interface software and technology}, pages 559--568. ACM, 2011.

\bibitem{li2015jointembedding}
Yangyan Li, Hao Su, Charles~Ruizhongtai Qi, Noa Fish, Daniel Cohen-Or, and
  Leonidas~J. Guibas.
\newblock Joint embeddings of shapes and images via cnn image purification.
\newblock {\em ACM Trans. Graph.}, 2015.

\bibitem{liu2015style}
Tianqiang Liu, Aaron Hertzmann, Wilmot Li, and Thomas Funkhouser.
\newblock Style compatibility for {3D} furniture models.
\newblock {\em ACM Transactions on Graphics (Proc. SIGGRAPH)}, 34(4), Aug.
  2015.

\bibitem{massa2016deep}
Francisco Massa, Bryan~C Russell, and Mathieu Aubry.
\newblock Deep exemplar 2d-3d detection by adapting from real to rendered
  views.
\newblock In {\em Proceedings of the IEEE Conference on Computer Vision and
  Pattern Recognition}, pages 6024--6033, 2016.

\bibitem{newcombe2011kinectfusion}
Richard~A Newcombe, Shahram Izadi, Otmar Hilliges, David Molyneaux, David Kim,
  Andrew~J Davison, Pushmeet Kohi, Jamie Shotton, Steve Hodges, and Andrew
  Fitzgibbon.
\newblock Kinectfusion: Real-time dense surface mapping and tracking.
\newblock In {\em Mixed and augmented reality (ISMAR), 2011 10th IEEE
  international symposium on}, pages 127--136. IEEE, 2011.

\bibitem{newell2016stacked}
Alejandro Newell, Kaiyu Yang, and Jia Deng.
\newblock Stacked hourglass networks for human pose estimation.
\newblock In {\em Computer Vision -- ECCV 2016}, pages 483--499. Springer
  International Publishing, 2016.

\bibitem{niessner2013hashing}
Matthias Nie{\ss}ner, Michael Zollh\"ofer, Shahram Izadi, and Marc Stamminger.
\newblock Real-time 3d reconstruction at scale using voxel hashing.
\newblock {\em ACM Transactions on Graphics (TOG)}, 2013.

\bibitem{ohbuchi2003shape}
Ryutarou Ohbuchi, Takahiro Minamitani, and Tsuyoshi Takei.
\newblock Shape-similarity search of 3d models by using enhanced shape
  functions.
\newblock In {\em Proceedings of Theory and Practice of Computer Graphics,
  2003.}, pages 97--104. IEEE, 2003.

\bibitem{osada2002shape}
Robert Osada, Thomas Funkhouser, Bernard Chazelle, and David Dobkin.
\newblock Shape distributions.
\newblock {\em ACM Transactions on Graphics (TOG)}, 21(4):807--832, 2002.

\bibitem{peng2015learning}
Xingchao Peng, Baochen Sun, Karim Ali, and Kate Saenko.
\newblock Learning deep object detectors from 3d models.
\newblock In {\em Proceedings of the IEEE International Conference on Computer
  Vision}, pages 1278--1286, 2015.

\bibitem{phamshrec2018}
Quang-Hieu Pham, Minh-Khoi Tran, Wenhui Li, Shu Xiang, Heyu Zhou, Weizhi Nie,
  Anan Liu, Yuting Su, Minh-Triet Tran, Ngoc-Minh Bui, et~al.
\newblock Shrec’18: Rgb-d object-to-cad retrieval.

\bibitem{qi2017pointnet}
Charles~R Qi, Hao Su, Kaichun Mo, and Leonidas~J Guibas.
\newblock Pointnet: Deep learning on point sets for 3d classification and
  segmentation.
\newblock {\em Proc. Computer Vision and Pattern Recognition (CVPR), IEEE},
  1(2):4, 2017.

\bibitem{qi2016volumetric}
Charles~Ruizhongtai Qi, Hao Su, Matthias Nie{\ss}ner, Angela Dai, Mengyuan Yan,
  and Leonidas Guibas.
\newblock Volumetric and multi-view cnns for object classification on 3d data.
\newblock In {\em Proc. Computer Vision and Pattern Recognition (CVPR), IEEE},
  2016.

\bibitem{rusu2009fast}
Radu~Bogdan Rusu, Nico Blodow, and Michael Beetz.
\newblock Fast point feature histograms (fpfh) for 3d registration.
\newblock In {\em Robotics and Automation, 2009. ICRA'09. IEEE International
  Conference on}, pages 3212--3217. Citeseer, 2009.

\bibitem{sundar2003skeleton}
Hari Sundar, Deborah Silver, Nikhil Gagvani, and Sven Dickinson.
\newblock Skeleton based shape matching and retrieval.
\newblock In {\em 2003 Shape Modeling International.}, pages 130--139. IEEE,
  2003.

\bibitem{tombari2010signature}
Federico Tombari, Samuele Salti, and Luigi Di~Stefano.
\newblock Unique signatures of histograms for local surface description.
\newblock In Kostas Daniilidis, Petros Maragos, and Nikos Paragios, editors,
  {\em Computer Vision -- ECCV 2010}, pages 356--369, Berlin, Heidelberg, 2010.
  Springer Berlin Heidelberg.

\bibitem{wang2018sgpn}
Weiyue Wang, Ronald Yu, Qiangui Huang, and Ulrich Neumann.
\newblock Sgpn: Similarity group proposal network for 3d point cloud instance
  segmentation.
\newblock In {\em Proceedings of the IEEE Conference on Computer Vision and
  Pattern Recognition}, pages 2569--2578, 2018.

\bibitem{weston2010large}
Jason Weston, Samy Bengio, and Nicolas Usunier.
\newblock Large scale image annotation: learning to rank with joint word-image
  embeddings.
\newblock {\em Machine learning}, 81(1):21--35, 2010.

\bibitem{weston2011wsabie}
Jason Weston, Samy Bengio, and Nicolas Usunier.
\newblock Wsabie: Scaling up to large vocabulary image annotation.
\newblock In {\em Twenty-Second International Joint Conference on Artificial
  Intelligence}, 2011.

\bibitem{whelan2015elasticfusion}
Thomas Whelan, Stefan Leutenegger, Renato~F Salas-Moreno, Ben Glocker, and
  Andrew~J Davison.
\newblock Elasticfusion: Dense slam without a pose graph.
\newblock {\em Proc. Robotics: Science and Systems, Rome, Italy}, 2015.

\end{thebibliography}
